\documentclass[conference]{IEEEtran}
\usepackage{stmaryrd}
\usepackage{amsfonts}
\usepackage{graphicx,times,amsmath}

\usepackage{amssymb}
\usepackage{algorithmic}
\usepackage{algorithm}
\usepackage{subfigure}
\usepackage{array}
\usepackage{multirow,tabularx}

\newcommand{\bfx}{{\textbf{x}}}

\newcommand{\bfw}{{\textbf{w}}}
\newcommand{\bfy}{{\textbf{y}}}

\newcommand{\bfb}{{\textbf{b}}}
\newcommand{\bfs}{{\textbf{s}}}

\IEEEoverridecommandlockouts

\begin{document}

\title{Semi-supervised sparse coding
\thanks{Jim Jing-Yan Wang and Xin Gao are with
Computer, Electrical and Mathematical Sciences and Engineering Division,
King Abdullah University of Science and Technology (KAUST),
Thuwal 23955-6900, Saudi Arabia.}
}

\author{Jim Jing-Yan Wang and Xin Gao}

\maketitle

\begin{abstract}
Sparse coding approximates the data sample as a sparse linear combination of some basic codewords and
uses the sparse codes as new presentations.
In this paper, we investigate
learning discriminative sparse codes by sparse coding in a semi-supervised manner, where only a few training samples
are labeled.
By using the manifold structure spanned by the data set of both labeled and unlabeled samples
and the constraints provided by the labels of the labeled samples,
we learn the variable class labels for all the samples. Furthermore, to improve the
discriminative ability of the learned sparse codes, we assume that
the class labels could be predicted from the sparse codes directly using a linear classifier.
By solving the codebook, sparse codes, class labels and classifier parameters simultaneously
in a unified objective function, we develop a semi-supervised sparse coding algorithm.
Experiments on two real-world pattern recognition problems demonstrate the advantage of the proposed methods over
supervised sparse coding methods on partially labeled data sets.
\end{abstract}

\section{Introduction}

\PARstart{S}{parse}  Coding (SC) \cite{lu2012heterogeneous,lu2015semantic,Lee2007801,lu2015noise,Liu6740901,zhou2014spatial,zhou2013sub,lu2011latent} has been a popular and effective
data representation method for many applications, including pattern recognition \cite{Liu2012176,lu2013learning,Jenatton20112297,wang2009image,lu2008unsupervised},
bioinformatics \cite{Chen2013,Zhang2012,Lei201335} and computer vision \cite{lu2011kernel,Wang20091643,lu2013learning,Yang20091794,lu2013exhaustive,lu2010combining,lu2009generalized,lu2011contextual}.
Given a data sample with its feature vector, SC tries to learn a codebook with some codeworks, and
approximate the data sample as the linear combination of the codewords.
SC assume that only a few codewords in the codebook are enough to represent the data sample,
thus the combination coefficients should be sparse, i.e. most of the coefficients are zeros,
leaving only a few of them non-zeros.
The linear combination coefficients of the data sample could be its new representation.
Because they are sparse, the coefficient vector is often referred to as the sparse code.
To solve the sparse code, one usually minimizes the approximation error with regard to the
codebook and the sparse code, and at the same time seeks the sparsity of the sparse code.

Although SC has been used in many pattern recognition applications, such as
palmprint recognition \cite{Shang2010112},
dynamic texture recognition \cite{Ghanem2010987},
human action recognition \cite{lu2013latent,Liu20123557,lu2011spectral},
speech recognition \cite{Sivaram20104346},
digit recognition  \cite{Labusch20081985},
image annotation \cite{lu2012image,lu2009context,lu2009image},
and face recognition \cite{Yang2011625}, in most cases,
SC is used as an unsupervised learning method.
When SC is performed to the training data set,
it is assumed that the class labels of the training samples are unavailable.
Then after the sparse codes are learned, they will be used to learn a classifier.
Thus the class labels are ignored during the sparse coding procedure.
However, in most pattern recognition problems, the class labels of the training samples are
given. It is thus natural to improve the discriminative ability of the learned sparse codes
for the classification purpose.
To solve this problem, a few supervised SC methods were proposed to include the class labels
during the coding of the samples.
For example, Mairal et al. \cite{Mairal20091033} proposed to learn the sparse codes of the samples and
a classifier in the sparse code space simultaneously, by constructing
and optimizing  a unified objective function for the
SC parameters and the classification parameters.
Wang et al. \cite{Wang2013199}
proposed the discriminative SC method based on multi-manifolds, by learning discriminative class-conditioned codebooks and sparse codes from both data feature spaces and class labels.
Though these methods use the class labels, they require that all the training samples are labeled.
However, in some real-world applications, there are only very few training samples labeled, while
the remaining training samples are unlabeled. Learning from such a training set is called semi-supervised learning \cite{Bilenko200481}.
Semi-supervised learning, compared to the supervised learning, can explore both
the labels of the labeled samples and the distribution of the overall data set containing labeled and unlabeled
samples.
When there are few labeled samples, they are not sufficient to learn an effective classifier using a supervised
learning algorithm. In this case, it is necessary to include the unlabeled samples to explore the overall
distribution.
Many semi-supervised learning algorithm has been proposed to learn classifier from both labeled and unlabeled samples
(inductive learning) \cite{Ching1995641},
or to learn the labels of the unlabeled samples from the labeled samples
(transductive learning) \cite{Yu20061081}.
However, surprisingly,  no work has been done to learn discriminate sparse codes
from partially labeled data set by utilizing both the labels and the feature vectors of the labeled samples, and the
feature vectors of the unlabeled data samples.
It is interesting to note that
He et al. \cite{He20112849} proposed to use the SC method to construct a sparse graph from the
data set for the transductive learning problem, so that the class labels could be prorogated from the labeled samples to the unlabeled samples
via the sparse code. However, during the sparse graph learning procedure using SC, the class
labels of the labeled samples were ignored. Thus in He et al.'s work \cite{He20112849}, SC was also performed in an unsupervised way.
Similarly, SC was also used to construct a sparse graph for the transductive learning problem \cite{lu2008semi,13722894}.

To fill this gap, we propose a semi-supervised SC method in this paper.
Given a data set with only few of the samples labeled, besides
conducting SC for all the samples, we also assume that the class labels for all the samples could be
learned from their sparse codes.
To do this, we define variable class labels for all the samples, and a classifier to predict the variable
class labels.
The variable class label learning is regularized by the manifold of the data set and the labels of the labeled samples.
To learn the codebook, sparse codes, variable class labels, and the classifier parameters simultaneously,
we propose a unified objective function. In the objective function, besides the approximation error term and the sparsity term for SC, we also introduce the class label  approximation error term and the manifold regularization term for variable class labels.
By optimizing this objective function, we try to predict the variable class label from the sparse codes, thus
the learned sparse code is naturally discriminative since it has the ability to predict the class labels.
Moreover, the learning of the class labels of the unlabeled samples is regularized by
the known labels of the labeled samples, the sparse codes and the manifold structure of the data set.
The contributions of this paper are in two folds:

\begin{enumerate}
\item We propose a discriminative SC method which could learn from semi-supervised data set. It is a
discriminative representation and both labeled and unlabeled data samples could be used to improve its discriminative power.
\item Moreover, it is also an inductive learning method since it learns a codebook and a classifier from the
semi-supervised training set, which could be further used to code and classify the test samples.
\end{enumerate}

The rest parts of this paper is organized as follows: in Section \ref{sec:method}, we introduce
the proposed semi-supervised SC method;
in Section \ref{sec:exp}, the experiment results on two data sets are reported;
and finally in Section \ref{sec:conclusion} the paper is concluded.

\section{Proposed Method}
\label{sec:method}

In this section, we introduce the proposed semi-supervised learning method.
An objective function is firstly constructed, and
then an iterative algorithm is developed to optimize it.

\subsection{Objective Function}

We assume that we have a training data set of $n$ training samples, denoted as
$\{\bfx_1,\cdots,\bfx_{n}\}\in \mathbb{R}^d$,
where $\bfx_i$ is the $d$-dimensional feature vector for the $i$-th sample.
The data set is further denoted as a data matrix as
$X=[\bfx_1,\cdots,\bfx_{n}]\in R^{d\times n}$, where the $i$-th column is the
feature vector of the $i$-th sample.
We assume that we are dealing with a $c$-class semi-supervised classification problem,
and only the first $l$ samples are labeled, while the remaining samples are unlabeled.
For a labeled sample $\bfx_i$, we define a $c$-dimensional binary class label vector
$\widehat{\bfy}_i\in \{1,0\}^{c}$, with its $\iota$-th element equal to one if it is labeled as the $\iota$-th class, and the reminding elements equal to zero.
The class label vector set of the labeled samples are denoted as
$\{\widehat{\bfy}_1,\cdots,\widehat{\bfy}_l\}\in \mathbb{R}^c$,
and they are further organized as a matrix
$\widehat{Y}_l=[\widehat{\bfy}_1,\cdots,\widehat{\bfy}_l]\in \{1,0\}^{c\times l}$, with its
$i$-th column as the label vector of the $i$-th sample.
To construct the objective function, we consider the following three problems:

\begin{itemize}
\item \textbf{Sparse Coding}:
Given a sample $\bfx_i$, sparse coding tries to learn a codebook matrix
$B=[\bfb_1,\cdots,\bfb_m]\in  \mathbb{R}^{d\times m}$, where its columns are $m$ codewords, and an $m$-dimensional coding vector $\bfs_i\in \mathbb{R}^m$, so that
$\bfx_i$ could be approximated as the linear combination of the codewords,

\begin{equation}
\begin{aligned}
\bfx_i \approx B \bfs_i
\end{aligned}
\end{equation}
And at the same time, $\bfs_i$ should be as sparse as possible. Thus we also call $\bfs_i$ sparse code.
The sparse code $\bfs_i$  is a new representation of $\bfx_i$.
The sparse codes of the training samples are organized in a sparse code matrix
$S=[\bfs_1,\cdots,\bfs_n]\in R^{m\times n}$, with its $i$-th column as the sparse code of the $i$-th sample.
To learn the codebook and the sparse codes from the training set, the following optimization problem is proposed,

\begin{equation}
\label{equ:sc}
\begin{aligned}
\underset{B,S}{\min}
~
&
\sum_{i=1}^n
\left\{
\|\bfx_i-B\bfs_i\|^2_2
+\alpha \|\bfs_i\|_1
\right\},
\\
s.t
~
&
\|\bfb_k\|_2^2 \leq c,
\end{aligned}
\end{equation}
where the first term $\|\bfx_i-B\bfs_i\|^2_2$ is the approximation error term, the second term $\|\bfs_i\|_1$ is introduced to
encourage the sparsity of each $\bfx_i$,
and $\alpha$ is a  trade-off parameter.
Moreover, $\|\bfb_k\|_2^2 \leq c$ is imposed to  to reduce the complexity of each codeword.

\item \textbf{Class Label Learning}: We also propose to learn the
class label vectors from the sparse code space for all the training samples by a linear function.
To do this, we introduce a variable label vector for each sample $\bfx_i$ as $\bfy_i\in \mathbb{R}^c$.
Please note that we relax it as a real value vector instead of a binary vector, and each element
presents its membership of each class.
The variable class label vector set for all the training samples are denoted as
$\{\bfy_1,\cdots,\bfy_n\}\in \mathbb{R}^c$, and further organized as a variable class label matrix,
$Y=[\bfy_1,\cdots,\bfy_n]\in \mathbb{R}^{c\times n}$.
We assume that its class label vector could be approximated from its sparse code by a linear classifier,

\begin{equation}
\begin{aligned}
\bfy_i \approx W \bfs_i,
\end{aligned}
\end{equation}
where $W\in \mathbb{R}^{c\times m}$ is the classifier parameter matrix.
To learn the class labels and the classifier parameter matrix, we propose the following
optimization problem,

\begin{equation}
\label{equ:label}
\begin{aligned}
\underset{S,W,Y}{\min}
~
&
\sum_{i=1}^n
\|\bfy_i - W \bfs_i\|_2^2
\\
s.t
~
&
\|\bfw_k\|_2^2 \leq e, k=1,\cdots,m\\
&
\bfy_i=\widehat{\bfy}_i, i=1,\cdots,l.
\end{aligned}
\end{equation}
As we can see from the above objective function,
we use the squared $L_2$ norm distance  $\|\bfy_i - W \bfs_i\|_2^2$
as the approximation error for the $i$-th sample.
Moreover, $\|\bfw_k\|_2^2 \leq e$ constrain is introduced to reduce the complexity of the classifier,
and $\bfy_i=\widehat{\bfy}_i, i=1,\cdots,l$ constrains are introduced so that
the learned labels could respect the known labels of the labeled samples.

\item \textbf{Manifold Label Regularization}:
We also hope the learned class labels could respect the manifold structure of the data set.
We assume that for each sample $\bfx_i$, its class label vector $\bfy_i$ could be
reconstructed by the class labels of its nearest neighbors $\mathcal{N}_i$,

\begin{equation}
\begin{aligned}
\bfy_i \approx \sum_{j\in \mathcal{N}_i}
A_{ij}
\bfy_j,
\end{aligned}
\end{equation}
where $A_{ij}$ is the reconstruction coefficient, which could be solved in the same way as
Locally Linear Embedding (LLE) \cite{roweis2000nonlinear}
by minimizing the reconstruction error in the original feature space,

\begin{equation}
\label{equ:QP}
\begin{aligned}
\underset{A_{ij}|_{j=1}^n}{\min}
~
&
\left\|
\bfx_i - \sum_{j\in \mathcal{N}_i}
A_{ij}
\bfx_j
\right\|^2_2
\\
s.t
~
&
A_{ij}\geq 0, j\in \mathcal{N}_i, ~\sum_{j\in \mathcal{N}_i}
A_{ij} = 1\\
&
A_{ij}=0 , j\notin \mathcal{N}_i
\end{aligned}
\end{equation}
With the solved reconstruction coefficient matrix $A=[A_{ij}]\in \mathbb{R}_+^{n\times n}$,
we regularize the class label learning with the following optimization problem,

\begin{equation}
\label{equ:manifold}
\begin{aligned}
\underset{Y}{\min}
~
&
\sum_{i=1}^n
\left\|
\bfy_i - \sum_{j\in \mathcal{N}_i}
A_{ij}
\bfy_j
\right\|^2_2
\\
s.t
~
&
\bfy_i=\widehat{\bfy}_i, i=1,\cdots,l.
\end{aligned}
\end{equation}
By doing this, we assume that
label space
and the data space share the same local linear reconstruction
coefficients.

\end{itemize}

The overall optimization problem is formulated by
combining the three problems in (\ref{equ:sc}), (\ref{equ:label}) and (\ref{equ:manifold}),
and the following optimization problem is obtained,

\begin{equation}
\label{equ:objective}
\begin{aligned}
\underset{B,S,Y,W}{\min}
~
&
\sum_{i=1}^n
\left\{
\|\bfx_i-B\bfs_i\|^2_2
+\alpha \|\bfs_i\|_1
+\beta\|\bfy_i - W \bfs_i\|_2^2
\vphantom{
\left\|
\bfy_i - \sum_{j\in \mathcal{N}_i}
A_{ij}
\bfy_j
\right\|^2_2
}
\right.\\
&
\left.
+
\gamma
\left\|
\bfy_i - \sum_{j\in \mathcal{N}_i}
A_{ij}
\bfy_j
\right\|^2_2
\right\}
\\
s.t.
~
&
\|\bfb_k\|_2^2 \leq c, \|\bfw_k\|_2^2 \leq e, k=1,\cdots,m,\\
&
\bfy_i=\widehat{\bfy}_i, i=1,\cdots,l.
\end{aligned}
\end{equation}
where $\beta$ and $\gamma$ are the tradeoff parameters, which are selected by cross-validation.
Please note that in this formulation, we do not use the
class labels to regularize the sparse codes directly. Instead,
a classifier is learned to assign the class label from the sparse codes,
so that the class labels, the classifiers, and the sparse codes could be
learned together and regularize each other.

\subsection{Optimization}

It is difficult to find a closed-form solution for the problem in (\ref{equ:objective}).
Thus we use the alternate  optimization strategy to optimize it in an iterative algorithm.
In each iteration,
the variables are optimized by turn.
When one of the variables is optimized, the others are fixed.

\subsubsection{Optimizing $B$ and $W$}

We first discuss the optimization of $B$ and $W$. As we show later, they could be solved together as different
parts of an generalized codebook.
By removing the terms irrelevant to $B$ and $W$, and fixing $S$ and $Y$, we obtain the following optimization problem,

\begin{equation}
\begin{aligned}
\underset{B,W}{\min}
~
&
\sum_{i=1}^n
\left\{
\|\bfx_i-B\bfs_i\|^2_2
+\beta\|\bfy_i - W \bfs_i\|_2^2
\right\}\\
&=
\left \| X-BS \right \|_2^2 + \left \|\sqrt{\beta}Y - \sqrt{\beta}W S\right \|_2^2
\\
s.t.
~
&
\|\bfb_k\|_2^2 \leq c, \|\bfw_k\|_2^2 \leq e, k=1,\cdots,m.
\end{aligned}
\end{equation}
We define an extended data matrix by catenating $X$ and $Y$ as $\widetilde{X}=
\begin{bmatrix}
X\\\sqrt{\beta}Y
\end{bmatrix}
$,
and an extended codebook matrix by catenating $B$ and $W$ as $\widetilde{B}=
\begin{bmatrix}
B\\\sqrt{\beta}W
\end{bmatrix}
$.
Moreover, we combine the two constrains $\|\bfb_k\|_2^2 \leq c$ and  $\|\bfw_k\|_2^2 \leq e$
to one single constraint $\|\bfb_k\|_2^2 + \beta \|\bfw_k\|_2^2 \leq c + \beta e$.
This constrain could be rewritten as
$\left\|
\begin{bmatrix}
\bfb_k\\\sqrt{\beta}\bfw_k
\end{bmatrix}
\right \|^2_2
= \|\widetilde{\bfb}_k\|_2^2
\leq (c+\beta e)
$,  where $\widetilde{\bfb}_k$ is the $k$-th column of the $\widetilde{B}$ matrix.
In this way, the optimization is rewritten as

\begin{equation}
\label{equ:B}
\begin{aligned}
\underset{\widetilde{B}}{\min}
&
\left\|
\widetilde{X}-\widetilde{B}S
\right\|^2_2
\\
s.t
&
\left\|
\widetilde{\bfb}_k
\right \|^2_2
\leq (c+\beta e), k=1,\cdots,m.
\end{aligned}
\end{equation}
This problem could be solved using the Lagrange dual method proposed in \cite{lee2006efficient}.
After $\widetilde{B}$ is solved, $B$ and $W$ could be recovered from it as

\begin{equation}
\begin{aligned}
&B=\widetilde{B}_{1,\cdots,d},\\
&W=\frac{1}{\sqrt{\beta}}\widetilde{B}_{d+1,\cdots,d+c},
\end{aligned}
\end{equation}
where $\widetilde{B}_{1,\cdots,d}$ is the frist $d$ rows of the matrix $\widetilde{B}$,
and $\widetilde{B}_{d+1,\cdots,d+c}$ is the $d+1$ to $d+c$ rows of matrix  $\widetilde{B}$.

\subsubsection{Optimizing $S$}

To solve the sparse codes in $S$, we fix $\widetilde{B}$, remove the terms irrelevant to $S$,
and the following problem is obtained,

\begin{equation}
\label{equ:S}
\begin{aligned}
\underset{\widetilde{B}}{\min}
&
\left\|
\widetilde{X}-\widetilde{B}S
\right\|^2_2
+\alpha
\sum_{i=1}^n \|\bfs_i\|_1
\end{aligned}
\end{equation}
Similarly, this problem could be solved efficiently by the
feature-sign search algorithm proposed in \cite{lee2006efficient}.

\subsubsection{Optimizing $Y$}

To solve the class label vectors in $Y$, we fix $B$, $S$ and $W$, remove the terms irrelevant to $Y$, and
get the following optimization problem,

\begin{equation}
\label{equ:obj_Y}
\begin{aligned}
\underset{Y}{\min}~
&\beta
\sum_{i=1}^n
\|\bfy_i - W \bfs_i\|_2^2
+
\gamma
\sum_{i=1}^n
\left\|
\bfy_i - \sum_{j\in \mathcal{N}_i}
A_{ij}
\bfy_j
\right\|^2_2
\\
&=
\beta\left \|
Y-WS
\right \|^2_2
+\gamma
\left \|
Y(I-A)^\top
\right \|_2^2\\
s.t~
&
\bfy_i=\widehat{\bfy}_i, i=1,\cdots,l.
\end{aligned}
\end{equation}
We separate the class label matrix to to sub-matrices as
$Y=[Y_l ~ Y_u]$, where $Y_l$ contains the first $l$ columns of $Y$, which are the variable class label vectors of the
labeled samples, while $Y_u$ contains the remaining columns which are the variable class label vectors of
the unlabeled samples. Similarly, we also separate $S$ to two sub-matrices as
$S=[S_l ~ S_u]$,
where $S_l$ contains the sparse codes of the labeled samples, while $S_u$ contains the sparse codes of the
labeled samples. Moreover, we define matrix $Q=(I-A)^\top$ for convenience, and also separate it to two
sub-matrices as
$
Q=
\begin{bmatrix}
Q_{l} \\
Q_{u}
\end{bmatrix}
$ where $Q_l$ contains its first $l$ rows and $Q_u$ contains its remaining rows.
With these definitions, we could rewrite the objective function in (\ref{equ:obj_Y}) as

\begin{equation}
\label{equ:obj_Y1}
\begin{aligned}
&\beta
\left \|
Y-WS
\right \|^2_2
+\gamma
\left \|
Y(I-A)^\top
\right \|_2^2\\
&=\beta
\left \|
Y_l-WS_l
\right \|^2_2
+\beta
\left \|
Y_u-WS_u
\right \|^2_2
+\gamma
\left \|
[Y_l ~Y_u]
\begin{bmatrix}
Q_{l} \\
Q_{u}
\end{bmatrix}
\right \|_2^2\\
&=\beta
\left \|
Y_l-WS_l
\right \|^2_2
+\beta
\left \|
Y_u-WS_u
\right \|^2_2
+\gamma
\left \|
Y_l Q_l + Y_u Q_u
\right \|_2^2
\end{aligned}
\end{equation}
Since it is constrained that $\bfy_i = \widehat{\bfy}_i$ for any $i=1,\cdots,l$,
$Y_l = \widehat{Y}_l$ and it is actually not a variable.
Thus we substitute $Y_l = \widehat{Y}_l$ to (\ref{equ:obj_Y1}) by only treating $Y_u$ as variable to solve,
and obtain the following optimization problem with regard to $Y_u$,

\begin{equation}
\begin{aligned}
\underset{Y_u}{\min}~
&
\left \{
f(Y_u)=
\beta
\left \|
\widehat{Y}_l-WS_l
\right \|^2_2
+
\beta
\left \|
Y_u-WS_u
\right \|^2_2
\right .\\
&
\left.
+\gamma
\left \|
\widehat{Y}_l Q_l + Y_u Q_u
\right \|_2^2
\right \}
\end{aligned}
\end{equation}
To solve this problem, we simply set the derivative of the objective function
$f(Y_u)$ with regard to $Y_u$ to zero, and obtain the solution for $Y_u$,

\begin{equation}
\label{equ:Y}
\begin{aligned}
&\frac{\partial f(Y_u)}{\partial Y_u}=
2 \beta\left ( Y_u-WS_u \right )+
2 \gamma  \left ( \widehat{Y}_l Q_l + Y_u Q_u \right ) Q_u^\top = 0\\
&\Rightarrow
Y_u =  \left (\beta WS_u - \gamma \widehat{Y}_l Q_l Q_u^\top \right )
\left ( \beta I + \gamma Q_u Q_u^\top \right )^{-1}
\end{aligned}
\end{equation}

\subsection{Algorithm}

We summarize the iterative learning algorithm for Semi-Supervised Sparse Coding (SSSC)
in Algorithm \ref{alg:learning}.
As we can see from the algorithm, we employ the original sparse coding algorithm
to initialize the sparse code matrix, and employ the
Linear Neighborhood Propagation (LNP) algorithm \cite{wang2009linear} to initialize the class label matrix.
The iterations are repeated for $T$ times and the updated solutions for $B$, $S$, $W$ and $Y_u$ are outputted.

\begin{algorithm}
\caption{Learning Algorithm of SSSC.}
\label{alg:learning}
\begin{algorithmic}
\STATE \textbf{Input}: Training data matrix $X$;
\STATE \textbf{Input}: Training data label matrix for labeled samples $\widehat{Y}_l$;
\STATE \textbf{Input}: Tradeoff parameters $\alpha$, $\beta$ and $\gamma$.;
\STATE \textbf{Input}: Iteration number $T$.

\STATE Initialize the sparse code matrix $S^0$ by performing original sparse coding to $X$;
\STATE Initialize the class label matrix $Y^0$;

\FOR{$t=1,\cdots,T$}

\STATE Update codebook matrix $B^t$ and the classifier parameter matrix $W^t$
as in (\ref{equ:B}) by fixing $S^{t-1}$ and $Y^{t-1}$;

\STATE Update sparse code matrix $S^t$
as in (\ref{equ:S}) by fixing $B^{t}$ and $Y^{t-1}$;

\STATE Update the variable class label matrix $Y^t$
as in (\ref{equ:Y}) by fixing $B^{t}$ and $S^{t}$;

\ENDFOR
\STATE \textbf{Output}: The codebook matrix $B^T$, the sparse code matrix $S^T$,
the classifier  parameter matrix $W^{T}$, and the
class label matrix for the unlabeled samples $Y_u^T$.
\end{algorithmic}
\end{algorithm}

\subsection{Coding and Classifying New Samples}

When a new test sample  $\bfx$ comes, we first find its nearest neighbors $\mathcal{N}$ from the
training set, and we assume that it could be reconstructed by these
nearest neighbors.
The  reconstruction coefficients $a_i|_{i\in \mathcal{N}}$ are computed by solving a problem in
(\ref{equ:QP}).
To solve its sparse code vector $\bfs$, and its class label vector $\bfy$, we use the codebook $B$,
classifier parameter matrix $W$, and the class label matrix $Y$  learned from
the training set.
The optimization problem is formulated as

\begin{equation}
\label{equ:objective_x}
\begin{aligned}
\underset{\bfs,\bfy}{\min}
~
&
\left\{
\|\bfx-B\bfs\|^2_2
+\alpha \|\bfs_i\|_1
+\beta\|\bfy - W \bfs\|_2^2
\vphantom{
\left\|
\bfy - \sum_{i\in \mathcal{N}}
a_{i}
\bfy_i
\right\|^2_2
}
\right.\\
&\left.
+
\gamma
\left\|
\bfy - \sum_{i\in \mathcal{N}}
a_{i}
\bfy_i
\right\|^2_2
\right\},
\end{aligned}
\end{equation}
where $\bfy_i$ is the class label vector of the $i$-th training sample.
To solve this problem, we also adopt the alternate  optimization strategy.
In an iterative algorithm, we optimize $\bfs$ and $\bfy$ in turn.
\begin{itemize}
\item\textbf{Solving $\bfs$}:
When $\bfs$ is optimized, $\bfy$ is fixed, and the following problem is solved,

\begin{equation}
\label{equ:s}
\begin{aligned}
\underset{\bfs}{\min}
~
&
\left\{
\|\bfx-B\bfs\|^2_2
+\alpha \|\bfs_i\|_1
+\beta\|\bfy - W \bfs\|_2^2
\right.\\
&
\left.
=
\|\widetilde{\bfx}-\widetilde{B}\bfs\|^2_2
+\alpha \|\bfs_i\|_1
\right\},
\end{aligned}
\end{equation}
where $\widetilde{\bfx}=
\begin{bmatrix}
\bfx\\
\sqrt{\beta} \bfy
\end{bmatrix}
$. This problem could be solved using the feature-sign search algorithm proposed in \cite{lee2006efficient}.

\item\textbf{Solving $\bfy$}: When $\bfs$ is fixed and $\bfy$ is optimized, we have the following problem,

\begin{equation}
\begin{aligned}
\underset{\bfy}{\min}
~
&
\left\{
\beta\|\bfy - W \bfs\|_2^2
+
\gamma
\left\|
\bfy - \sum_{i\in \mathcal{N}}
a_{i}
\bfy_i
\right\|^2_2
\right\}.
\end{aligned}
\end{equation}
It could be solved easily by setting the derivative with regard to $\bfy$ to zero, and the solution is obtained as

\begin{equation}
\begin{aligned}
\bfy = \frac{1}{\beta+\gamma}
\left (
\beta W\bfs+ \gamma \sum_{i\in \mathcal{N}} a_i \bfy_i
\right )
\end{aligned}
\end{equation}

\end{itemize}
By repeating the above two procedures for $T$ times, we could obtain the optimal
sparse code $\bfs$ and the class label vector $\bfy$ for the test sample $\bfx$.
It will be further classifier to the $\iota^*$-th class with  the largest value in
the class label vector $\bfy$,

\begin{equation}
\begin{aligned}
\iota^* = {\arg\max}_{\iota\in \{1,\cdots,c\}} \bfy(\iota),
\end{aligned}
\end{equation}
where $\bfy(\iota)$ is the $\iota$-th element of $\bfy$.

\section{Experiments}
\label{sec:exp}

In this section, we evaluate the performance of the proposed semi-supervised sparse coding
algorithm on two real-world data sets.

\subsection{Cytochromes P450 Inhibition Prediction}

The cytochromes P450 is a family of enzymes which are involved in the metabolism of
most modern drugs \cite{Simpson1994342,BajRossi2014283,Rasmussen2014255}.
There are five major isoforms of cytochromes P450, which are
1A2, 2C9, 2C19, 2D6, and 3A4 \cite{Guengerich2006E105}.
It is very important to model the interactions of the  cytochromes P450  with
the drug-like compounds in drug-drug interaction studies.
In this case, predicting if a given compound  can inhibit
these isoforms plays an important role in the drug design \cite{Rostkowski20132051}.
Here, we evaluated the proposed algorithm in the problem of
cytochromes P450 inhibition prediction.

\subsubsection{Data Set and Protocol}

\begin{figure}[!t]
\centering
\includegraphics[width=0.5\textwidth]{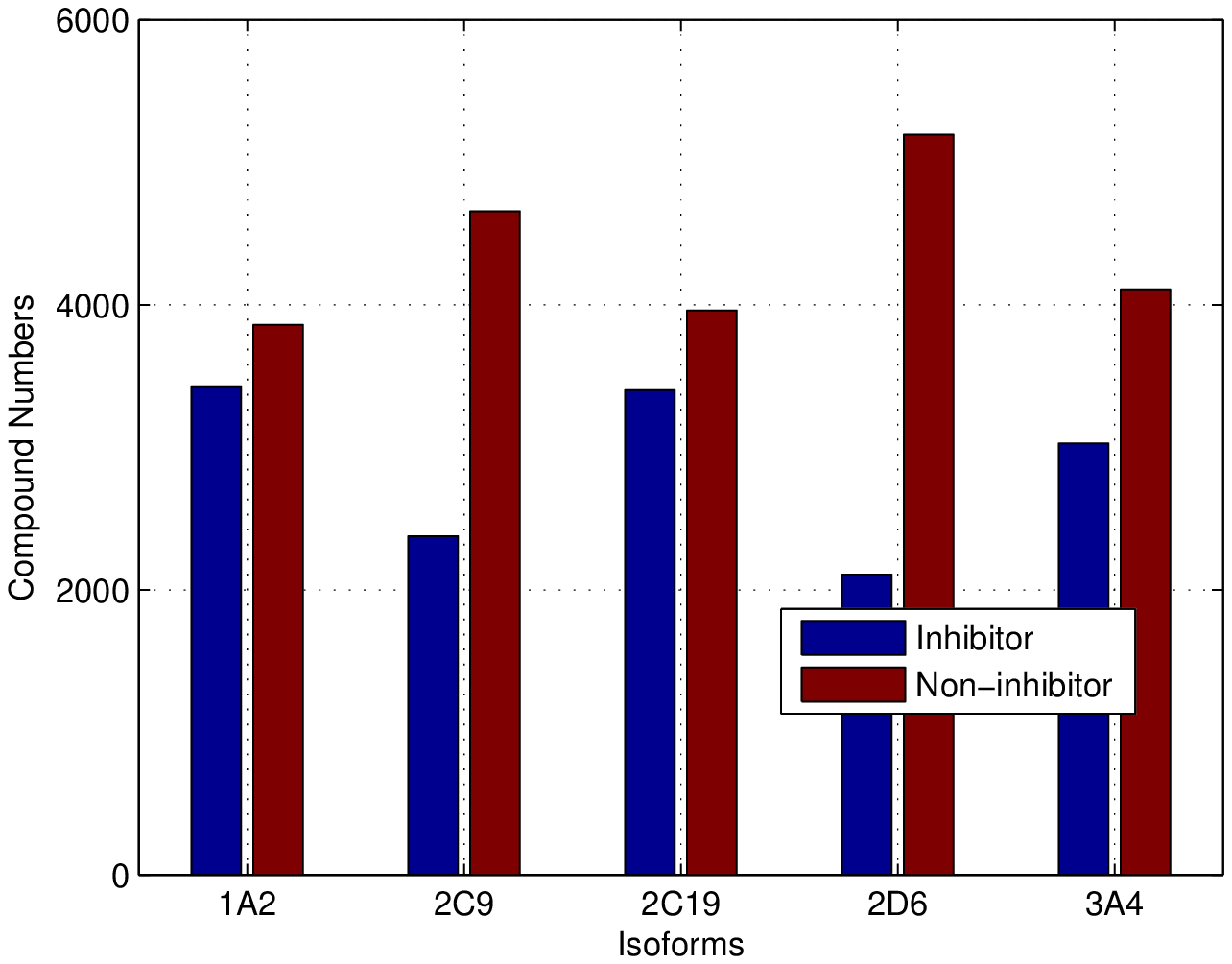}\\
\caption{The numbers of inhibitors and non-inhibitors of each isoform in the
cytochromes P450 inhibition prediction data set.}
\label{fig:FigP450_Data}
\end{figure}

We collected a data set of compounds for each isoform,
and each  compound is an inhibitor or a non-inhibitor of the isoform.
The numbers of inhibitors and non-inhibitors of each isoform are given in
Figure \ref{fig:FigP450_Data}.
As we can see from the figure, the data sets are not balanced. For each isoform, non-inhibitors are usually more
than inhibitors.
To represent each compound, we extracted the molecular signatures
as features,
which were computed from the
atomic signatures  of circular atomic fragments \cite{Faulon2008225,Faulon2003721,Faulon2003707}.
The problem of cytochromes P450 inhibition prediction is to learn a predictor from the
given data set to predict whether a candidate compound is an inhibitor or a non-inhibitor.
Thus it is a binary classification problem.

To conduct the experiment, for each isoform, we performed the 10-fold cross-validation \cite{Rojatkar20131222} to the
data set.
Each data set of an isoform was split into ten folds, and each fold was used as the test set in turn, while the
remaining nine folds were used as the training set.
For each taining set, we only randomly labeled a small part (about 20\%)of the compounds with the class labels (inhibitors or non-inhibitors),
while leaving the remaining part as unlabeled compounds.
The proposed learning algorithm was performed to the molecular signatures of the training compounds to learn
the codebook, the classifier and the labels of the unlabeled compounds.
Then the compounds in the test set were used as test sample one by one.
The learned codebook and the classifier were used to code and classify the test compound.

To evaluate the prediction performance, we used the following performance measures as
prediction performance metrics: Sensitivity (Sen), Specificity (Spc), Accuracy (Acc),
and
F1 score (F1).
To calculate these metrics, we first calculate the following values for each test set:
True Positive (TP) which is the number of inhibitor compounds that were correctly predicted,
True Negative (TN) which is the number of non-inhibitor compounds that were correctly predicted,
False Positive (FP) which is the number of non-inhibitor compounds wrongly predicted as inhibitor compounds,
and
False Negative (FN) which is the number of inhibitor compounds wrongly predicted as non-inhibitor compounds.
With these values computed from the
test set, the performance measures are defined as,

\begin{equation}
\begin{aligned}
Sen&=\frac{TP}{TP+FN}, Spc=\frac{TN}{FP+TN},\\
Acc&=\frac{TP+TN}{TP+TN+FP+FN},\\
F1&=\frac{2\times TP}{2\times TP +  FP + FN }.
\end{aligned}
\end{equation}
Please note that the ranges of Sen, Spc, Acc and F1 values are all from $0$ to $1$, and a
larger value indicates a better prediction performance.

\subsubsection{Results}

\begin{figure}[!t]
\centering
\includegraphics[width=0.5\textwidth]{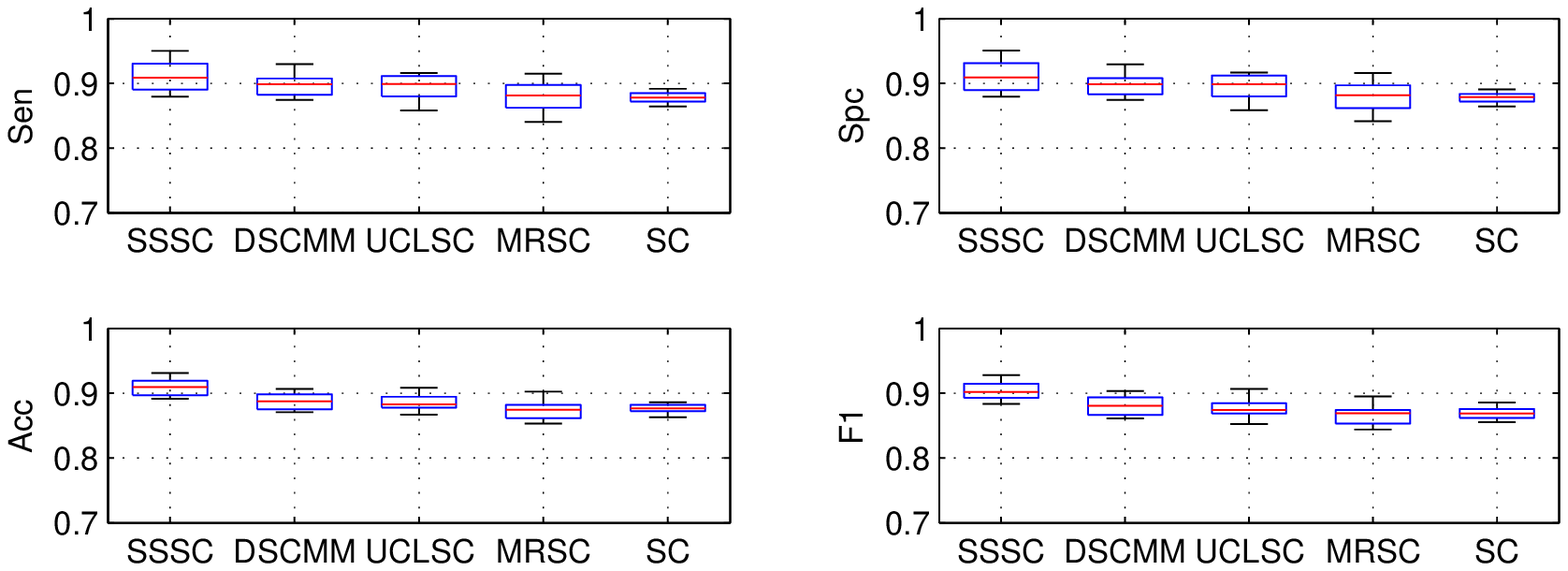}
\\
\caption{Experimental results on the 1A2 inhibitor data set.}
\label{fig:Fig1A2}
\end{figure}

\begin{figure}[!t]
\centering
\includegraphics[width=0.5\textwidth]{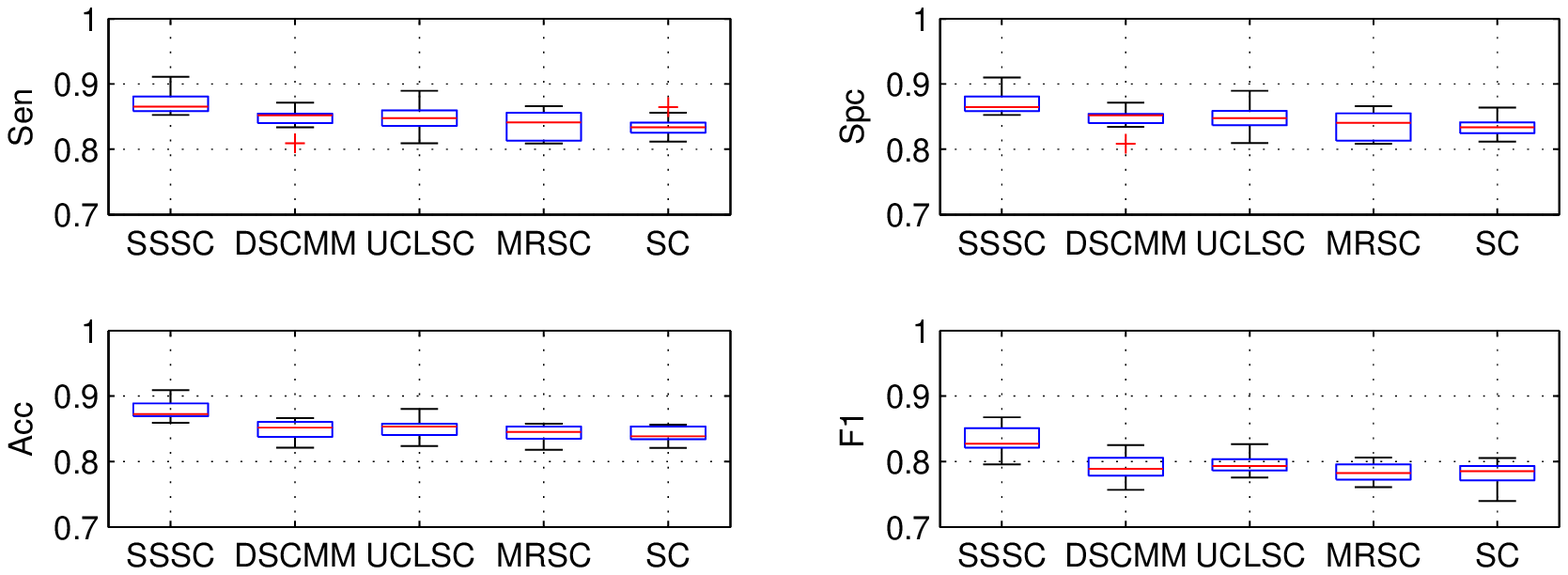}
\\
\caption{Experimental results on the 2C9 inhibitor data set.}
\label{fig:Fig2C9}
\end{figure}

\begin{figure}[!t]
\centering
\includegraphics[width=0.5\textwidth]{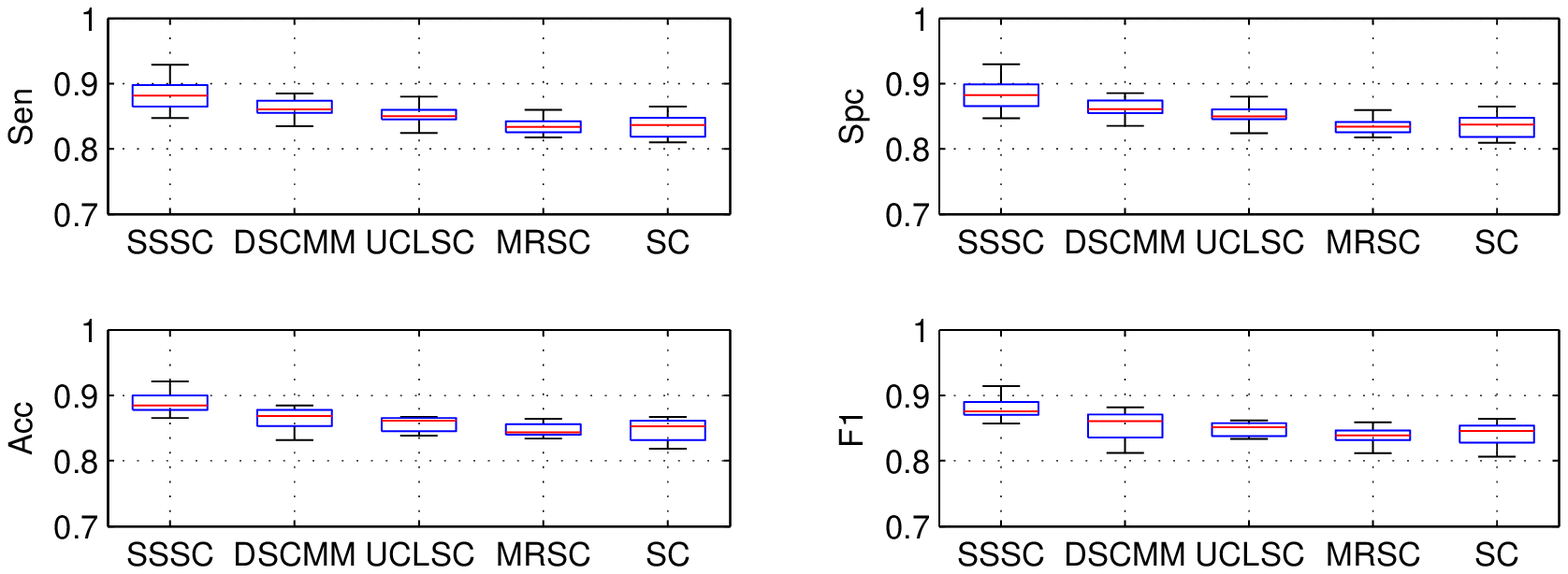}
\\
\caption{Experimental results on the 2C19 inhibitor data set.}
\label{fig:Fig2C19}
\end{figure}

\begin{figure}[!t]
\centering
\includegraphics[width=0.5\textwidth]{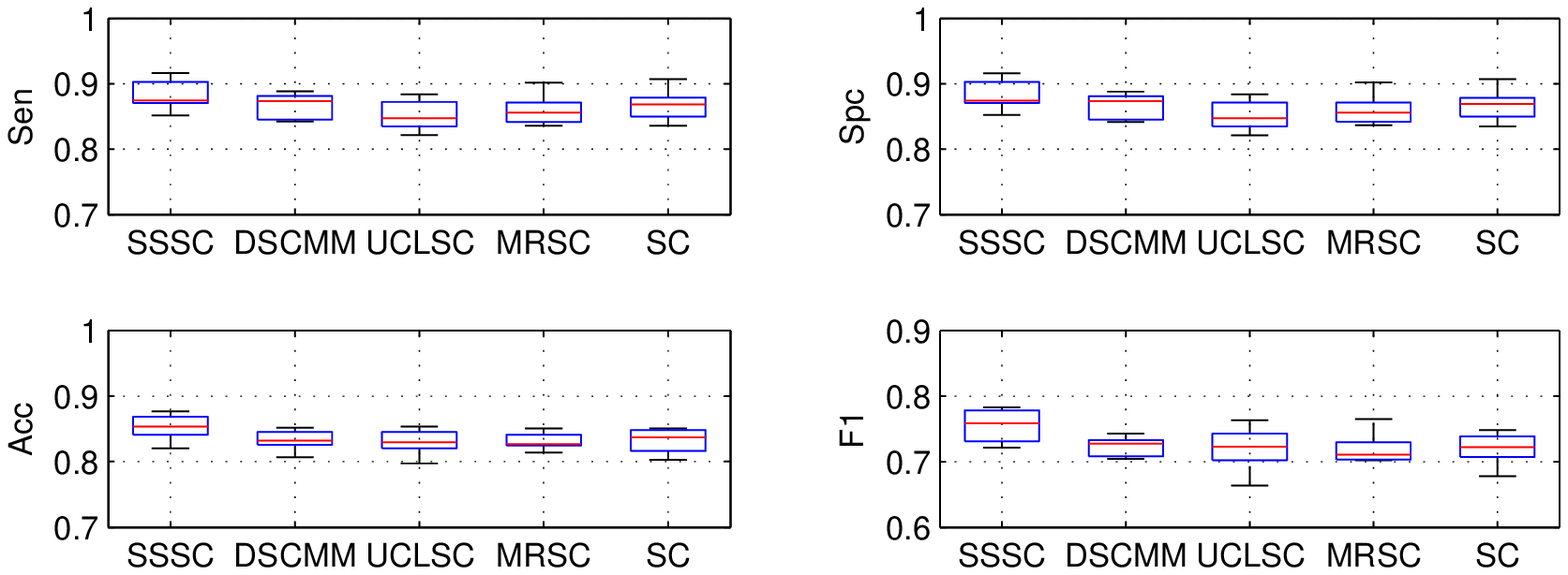}
\\
\caption{Experimental results on the 2D6 inhibitor data set.}
\label{fig:Fig2D6}
\end{figure}

\begin{figure}[!t]
\centering
\includegraphics[width=0.5\textwidth]{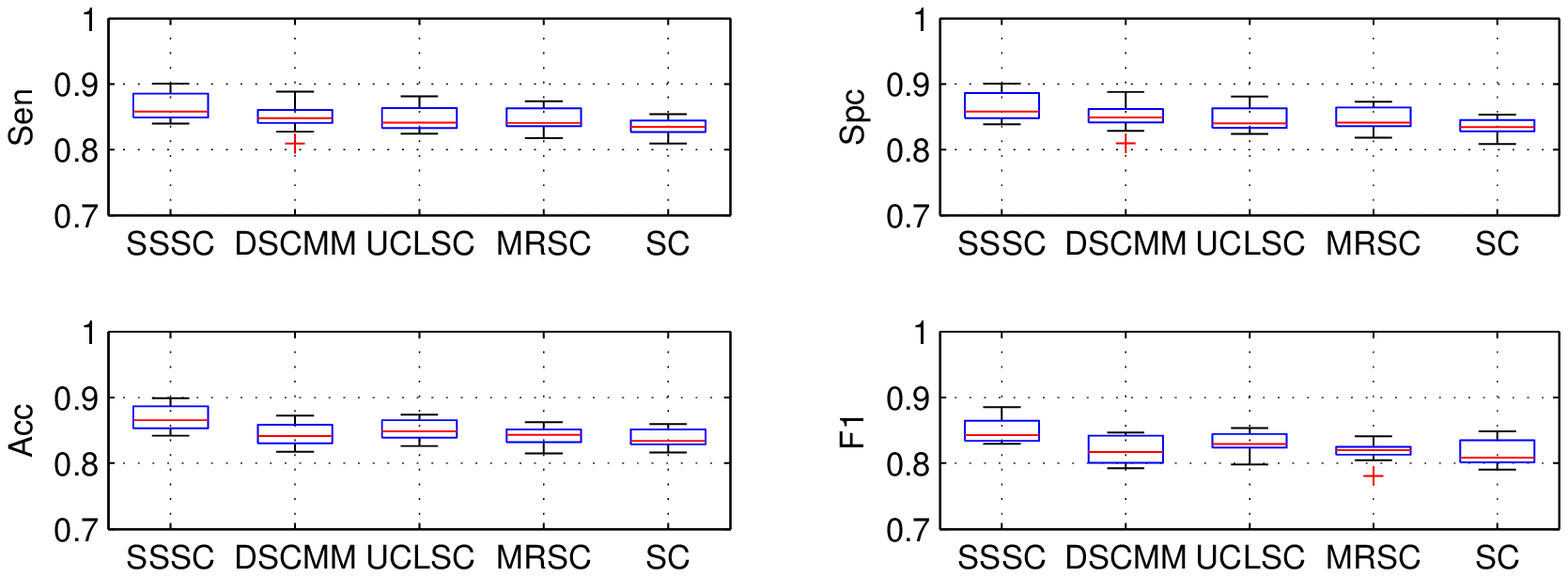}
\\
\caption{Experimental results on the 3A4 inhibitor data set.}
\label{fig:Fig3A4}
\end{figure}

Since the proposed algorithm is the first semi-supervised sparse coding algorithm,
we compared it to some unsupervised and supervised sparse coding algorithms.
For the unsupervised sparse coding algorithms, we compared the proposed SSSC against the original
sparse coding  (SC) algorithm proposed in \cite{Lee2007801},
and the popular manifold regularized sparse coding (MRSC) algorithm proposed in \cite{gao2010local}.
For the supervised sparse coding algorithm, we compared it against the
unified classifier learning and sparse coding  (UCLSC) algorithm proposed in \cite{Mairal20091033},
and the discriminative sparse coding on multi-manifold (DSCMM) algorithm proposed in \cite{Wang2013199}.
Please note that for the supervised sparse coding algorithms, it is required that all the
training samples are labeled. In this case, we only used the labeled samples in the training set, while
the unlabeled samples were ignored.
The experiment results of four different performance measures on the
five data sets are given in Fig. \ref{fig:Fig1A2} - \ref{fig:Fig3A4}.
It is clear that our SSSC algorithm consistently outperforms
all other supervised and unsupervised sparse coding algorithms, namely
DSCMM, UCLSC, MRSC and SC, in terms of the
Sen, Spc, Acc and F1 measures. This implies that SSSC is able to learn more discriminative
sparse codes to distinguish  inhibitors from  non-inhibitors
by learning discriminative codebooks and classifiers.
The performance of
supervised methods, DSCMM and UCLSC, is comparable to that of unsupervised methods, MRSC and SC.
We should note that only labels are used by the supervised sparse coding methods, while
unsupervised methods can explore all samples.
However, supervised methods include class labels to improve the discriminative ability of the sparse codes
during learning, but unsupervised methods simply ignore them.
Only the proposed semi-supervised method, SSSC, can use both the labels and all samples.
Thus it is not surprising that it archives the best performance.

\subsection{Wireless Sensor Fault Diagnosis}

In this experiment, we evaluate the proposed algorithm on the problem of
wireless sensor fault diagnosis for wireless networks \cite{DePaola2013}.

\subsubsection{Data Set and Setup}

We collected a data set of 300 samples of wireless sensors.
The samples were classified to
four fault types, including
shock, biasing, short circuit, and shifting. We also included the normal type,
making it five types in total. For each type, there are 60 samples.
For each sample, we used the output signal of wireless sensors as the feature to predict its
state type.

To conduct the experiment, we also employed the 10-fold cross validation. The entire data set was
split to 10 folds randomly. Each fold was used as the test set in turn, and the remaining nine folds
were combined and used as the training set to train the diagnosis model.
Most of the training samples were unlabeled while only a small portion of the training samples
was labeled. We performed the proposed algorithm to learn the
codebook, classifier, sparse codes and class labels of the unlabeled training samples.
The learned codebook and classifier are used to represent and classify the
test samples.
The classification performance is measured by the classification accuracy (Acc)
for multi-class problem, which is defined as follows,

\begin{equation}
\label{equ:acc}
\begin{aligned}
Acc=\frac{Number~of~correctly~classified~test~samples}{Number~of~test~samples}
\end{aligned}
\end{equation}
The value of Acc also varies from 0 to 1, and a larger Acc indicates  better classification
performance.

\subsubsection{Results}

\begin{figure}[!ttb]
\centering
\includegraphics[width=0.5\textwidth]{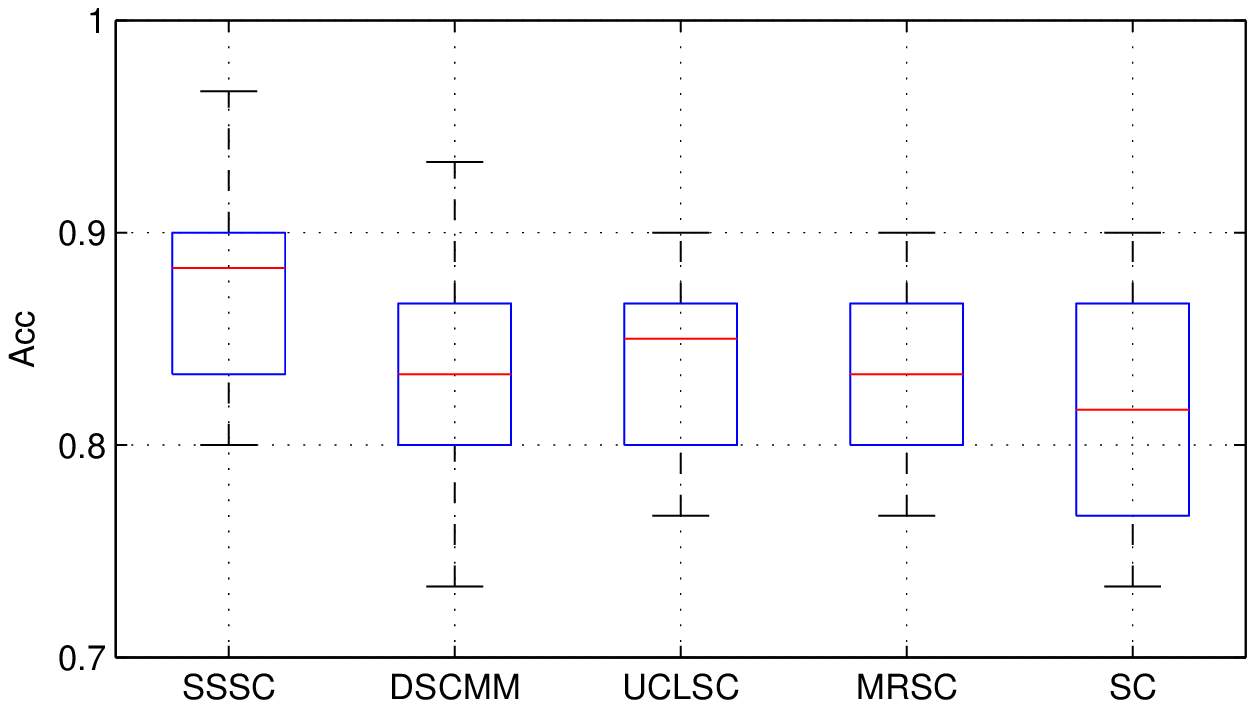}\\
\caption{Experimental results on the wireless sensor fault diagnosis data set.}
\label{fig:FigWireless131219}
\end{figure}

The boxplots of the accuracy of 10-fold cross validations are given in Fig. \ref{fig:FigWireless131219}.
From this figure, we can see that the proposed semi-supervised sparse coding and classification method SSSC
significantly outperforms the other sparse coding methods on the wireless sensor
fault diagnosis task.
This is because our method utilizes both the labeled and unlabeled samples
in learning the sparse code,
while others do not effectively use such
information.
Again, the supervised methods DSCMM and UCLSC do not show much better improvement
over the unsupervised methods MRSC and SC.
It is clear that the proposed SSSC combines the advantages of both supervised and unsupervised
methods. The codebook and the class labels of unlabeled samples are
directly learned from training samples.
Thus it is better adaptive to the data and higher classification accuracy can be achieved.

\section{Conclusion}
\label{sec:conclusion}

We have proposed a sparse coding
method for the semi-supervised data representation and classification task. To the best of our knowledge,
this paper is the first attempt to learn sparse code on partially labeled data sets.
Experimental
results have shown that our proposed method SSSC
are not only significantly better than state-of-the-art
unsupervised sparse coding methods, but also outperforms
supervised sparse coding methods.
How to explore more discriminative information from both labeled and unlabeled,
and combine them with our proposed
semi-supervised sparse coding algorithm to
further improve the learning
performance appears to be an interesting direction
in machine learning and pattern recognition communities.
In the future, we will investigate the usage of the proposed method in applications of bioinformatics \cite{tian14a,tian14b}.



\begin{thebibliography}{10}
\providecommand{\url}[1]{#1}
\csname url@samestyle\endcsname
\providecommand{\newblock}{\relax}
\providecommand{\bibinfo}[2]{#2}
\providecommand{\BIBentrySTDinterwordspacing}{\spaceskip=0pt\relax}
\providecommand{\BIBentryALTinterwordstretchfactor}{4}
\providecommand{\BIBentryALTinterwordspacing}{\spaceskip=\fontdimen2\font plus
\BIBentryALTinterwordstretchfactor\fontdimen3\font minus
  \fontdimen4\font\relax}
\providecommand{\BIBforeignlanguage}[2]{{%
\expandafter\ifx\csname l@#1\endcsname\relax
\typeout{** WARNING: IEEEtran.bst: No hyphenation pattern has been}%
\typeout{** loaded for the language `#1'. Using the pattern for}%
\typeout{** the default language instead.}%
\else
\language=\csname l@#1\endcsname
\fi
#2}}
\providecommand{\BIBdecl}{\relax}
\BIBdecl

\bibitem{lu2012heterogeneous}
Z.~Lu and Y.~Peng, ``Heterogeneous constraint propagation with constrained
  sparse representation,'' in \emph{Proceedings of IEEE International
  Conference on Data Mining}.\hskip 1em plus 0.5em minus 0.4em\relax IEEE
  Computer Society, 2012, pp. 1002--1007.

\bibitem{lu2015semantic}
Z.~Lu, P.~Han, L.~Wang, and J.-R. Wen, ``Semantic sparse recoding of visual
  content for image applications,'' \emph{IEEE Transactions on Image
  Processing}, vol.~24, no.~1, 2015.

\bibitem{Lee2007801}
H.~Lee, A.~Battle, R.~Raina, and A.~Ng, ``Efficient sparse coding algorithms,''
  in \emph{Advances in Neural Information Processing Systems}, 2007, pp.
  801--808.

\bibitem{lu2015noise}
Z.~Lu and L.~Wang, ``Noise-robust semi-supervised learning via fast sparse
  coding,'' \emph{Pattern Recognition}, vol.~48, no.~2, pp. 605--612, 2015.

\bibitem{Liu6740901}
L.~Liu, M.~Esmalifalak, Q.~Ding, V.~Emesih, and Z.~Han, ``Detecting false data
  injection attacks on power grid by sparse optimization,'' \emph{IEEE
  Transactions on Smart Grid}, vol.~5, no.~2, pp. 612--621, March 2014.

\bibitem{zhou2014spatial}
L.~Zhou, Z.~Lu, H.~Leung, and L.~Shang, ``Spatial temporal pyramid matching
  using temporal sparse representation for human motion retrieval,'' \emph{The
  Visual Computer}, vol.~30, no. 6-8, pp. 845--854, 2014.

\bibitem{zhou2013sub}
G.~Zhou, Z.~Lu, and Y.~Peng, ``$\ell_1$-graph construction using structured
  sparsity,'' \emph{Neurocomputing}, vol. 120, pp. 441--452, 2013.

\bibitem{lu2011latent}
Z.~Lu and Y.~Peng, ``Latent semantic learning by efficient sparse coding with
  hypergraph regularization.'' in \emph{AAAI}, 2011.

\bibitem{Liu2012176}
S.~Liu and M.~Liu, ``Fingerprint orientation modeling by sparse coding,'' in
  \emph{Proceedings - 2012 5th IAPR International Conference on Biometrics, ICB
  2012}, 2012, pp. 176--181.

\bibitem{lu2013learning}
Z.~Lu and Y.~Peng, ``Learning descriptive visual representation by semantic
  regularized matrix factorization,'' in \emph{IJCAI}.\hskip 1em plus 0.5em
  minus 0.4em\relax AAAI Press, 2013, pp. 1523--1529.

\bibitem{Jenatton20112297}
R.~Jenatton, J.~Mairal, G.~Obozinski, and F.~Bach, ``Proximal methods for
  hierarchical sparse coding,'' \emph{Journal of Machine Learning Research},
  vol.~12, pp. 2297--2334, 2011.

\bibitem{wang2009image}
L.~Wang, Z.~Lu, and H.~H. Ip, ``Image categorization based on a hierarchical
  spatial markov model,'' in \emph{International Conference on Computer
  Analysis of Images and Patterns}.\hskip 1em plus 0.5em minus 0.4em\relax
  Springer Berlin Heidelberg, 2009, pp. 766--773.

\bibitem{lu2008unsupervised}
Z.~Lu, Y.~Peng, and J.~Xiao, ``Unsupervised learning of finite mixtures using
  entropy regularization and its application to image segmentation,'' in
  \emph{IEEE Conference on Computer Vision and Pattern Recognition}.\hskip 1em
  plus 0.5em minus 0.4em\relax IEEE, 2008, pp. 1--8.

\bibitem{Chen2013}
S.~Chen, C.-Y. Zhang, and K.~Song, ``Recognizing short coding sequences of
  prokaryotic genome using a novel iteratively adaptive sparse partial least
  squares algorithm,'' \emph{Biology Direct}, vol.~8, no.~1, 2013.

\bibitem{Zhang2012}
K.~Zhang, J.~Han, T.~Groesser, G.~Fontenay, and B.~Parvin, ``Inference of
  causal networks from time-varying transcriptome data via sparse coding,''
  \emph{PLoS ONE}, vol.~7, no.~8, 2012.

\bibitem{Lei201335}
Z.~Lei, K.~Chen, H.~Li, H.~Liu, and A.~Guo, ``The gaba system regulates the
  sparse coding of odors in the mushroom bodies of drosophila,''
  \emph{Biochemical and Biophysical Research Communications}, vol. 436, no.~1,
  pp. 35--40, 2013.

\bibitem{lu2011kernel}
Z.~Lu and L.~Wang, ``Learning descriptive visual representation for image
  classification and annotation,'' \emph{Pattern Recognition}, vol.~48, no.~2,
  pp. 498--508, 2015.

\bibitem{Wang20091643}
C.~Wang, S.~Yan, L.~Zhang, and H.-J. Zhang, ``Multi-label sparse coding for
  automatic image annotation,'' in \emph{2009 IEEE Computer Society Conference
  on Computer Vision and Pattern Recognition Workshops, CVPR Workshops 2009},
  2009, pp. 1643--1650.

\bibitem{Yang20091794}
J.~Yang, K.~Yu, Y.~Gong, and T.~Huang, ``Linear spatial pyramid matching using
  sparse coding for image classification,'' in \emph{2009 IEEE Computer Society
  Conference on Computer Vision and Pattern Recognition Workshops, CVPR
  Workshops 2009}, 2009, pp. 1794--1801.

\bibitem{lu2013exhaustive}
Z.~Lu and Y.~Peng, ``Exhaustive and efficient constraint propagation: A
  graph-based learning approach and its applications,'' \emph{International
  Journal of Computer Vision}, vol. 103, no.~3, pp. 306--325, 2013.

\bibitem{lu2010combining}
Z.~Lu and H.~H.-S. Ip, ``Combining context, consistency, and diversity cues for
  interactive image categorization,'' \emph{IEEE Transactions on Multimedia},
  vol.~12, no.~3, pp. 194--203, 2010.

\bibitem{lu2009generalized}
Z.~Lu and H.~H. Ip, ``Generalized relevance models for automatic image
  annotation,'' in \emph{Advances in Multimedia Information Processing-PCM
  2009}.\hskip 1em plus 0.5em minus 0.4em\relax Springer Berlin Heidelberg,
  2009, pp. 245--255.

\bibitem{lu2011contextual}
Z.~Lu, H.~H. Ip, and Y.~Peng, ``Contextual kernel and spectral methods for
  learning the semantics of images,'' \emph{IEEE Transactions on Image
  Processing}, vol.~20, no.~6, pp. 1739--1750, 2011.

\bibitem{Shang2010112}
L.~Shang, W.~Huai, G.~Dai, J.~Chen, and J.~Du, ``Palmprint recognition using
  2d-gabor wavelet based sparse coding and rbpnn classifier,'' \emph{Lecture
  Notes in Computer Science (including subseries Lecture Notes in Artificial
  Intelligence and Lecture Notes in Bioinformatics)}, vol. 6064 LNCS, no. PART
  2, pp. 112--119, 2010.

\bibitem{Ghanem2010987}
B.~Ghanem and N.~Ahuja, ``Sparse coding of linear dynamical systems with an
  application to dynamic texture recognition,'' in \emph{Proceedings -
  International Conference on Pattern Recognition}, 2010, pp. 987--990.

\bibitem{lu2013latent}
Z.~Lu and Y.~Peng, ``Latent semantic learning with structured sparse
  representation for human action recognition,'' \emph{Pattern Recognition},
  vol.~46, no.~7, pp. 1799--1809, 2013.

\bibitem{Liu20123557}
Y.~Liu and Y.~Li, ``Human action recognition in videos using distance image
  volumes and sparse coding,'' \emph{Journal of Computational Information
  Systems}, vol.~8, no.~9, pp. 3557--3564, 2012.

\bibitem{lu2011spectral}
Z.~Lu, Y.~Peng, and H.~H.-S. Ip, ``Spectral learning of latent semantics for
  action recognition,'' in \emph{IEEE International Conference onComputer
  Vision (ICCV)}.\hskip 1em plus 0.5em minus 0.4em\relax IEEE, 2011, pp.
  1503--1510.

\bibitem{Sivaram20104346}
G.~Sivaram, S.~Nemala, M.~Elhilali, T.~Tran, and H.~Hermansky, ``Sparse coding
  for speech recognition,'' in \emph{ICASSP, IEEE International Conference on
  Acoustics, Speech and Signal Processing - Proceedings}, 2010, pp. 4346--4349.

\bibitem{Labusch20081985}
K.~Labusch, E.~Barth, and T.~Martinetz, ``Simple method for high-performance
  digit recognition based on sparse coding,'' \emph{IEEE Transactions on Neural
  Networks}, vol.~19, no.~11, pp. 1985--1989, 2008.

\bibitem{lu2012image}
Z.~Lu and Y.~Peng, ``Image annotation by semantic sparse recoding of visual
  content,'' in \emph{Proceedings of the 20th ACM international conference on
  Multimedia}.\hskip 1em plus 0.5em minus 0.4em\relax ACM, 2012, pp. 499--508.

\bibitem{lu2009context}
Z.~Lu, H.~H. Ip, and Q.~He, ``Context-based multi-label image annotation,'' in
  \emph{Proceedings of the ACM International Conference on Image and Video
  Retrieval}.\hskip 1em plus 0.5em minus 0.4em\relax ACM, 2009, p.~30.

\bibitem{lu2009image}
Z.~Lu and H.~H.-S. Ip, ``Image categorization with spatial mismatch kernels,''
  in \emph{IEEE Conference on Computer Vision and Pattern Recognition}.\hskip
  1em plus 0.5em minus 0.4em\relax IEEE, 2009, pp. 397--404.

\bibitem{Yang2011625}
M.~Yang, L.~Zhang, J.~Yang, and D.~Zhang, ``Robust sparse coding for face
  recognition,'' in \emph{Proceedings of the IEEE Computer Society Conference
  on Computer Vision and Pattern Recognition}, 2011, pp. 625--632.

\bibitem{Mairal20091033}
J.~Mairal, F.~Bach, J.~Ponce, G.~Sapiro, and A.~Zisserman, ``Supervised
  dictionary learning,'' in \emph{Advances in Neural Information Processing
  Systems 21 - Proceedings of the 2008 Conference}, 2009, pp. 1033--1040.

\bibitem{Wang2013199}
J.~J.-Y. Wang, H.~Bensmail, N.~Yao, and X.~Gao, ``Discriminative sparse coding
  on multi-manifolds,'' \emph{Knowledge-Based Systems}, vol.~54, no.~0, pp. 199
  -- 206, 2013.

\bibitem{Bilenko200481}
M.~Bilenko, S.~Basu, and R.~Mooney, ``Integrating constraints and metric
  learning in semi-supervised clustering,'' in \emph{Proceedings, Twenty-First
  International Conference on Machine Learning, ICML 2004}, 2004, pp. 81--88.

\bibitem{Ching1995641}
J.~Y. Ching, A.~K. Wong, and K.~C. Chan, ``Class-dependent discretization for
  inductive learning from continuous and mixed-mode data,'' \emph{IEEE
  Transactions on Pattern Analysis and Machine Intelligence}, vol.~17, no.~7,
  pp. 641--651, 1995.

\bibitem{Yu20061081}
K.~Yu, J.~Bi, and V.~Tresp, ``Active learning via transductive experimental
  design,'' in \emph{ICML 2006 - Proceedings of the 23rd International
  Conference on Machine Learning}, vol. 2006, 2006, pp. 1081--1088.

\bibitem{He20112849}
R.~He, W.-S. Zheng, B.-G. Hu, and X.-W. Kong, ``Nonnegative sparse coding for
  discriminative semi-supervised learning,'' in \emph{Proceedings of the IEEE
  Computer Society Conference on Computer Vision and Pattern Recognition},
  2011, pp. 2849--2856.

\bibitem{lu2008semi}
Z.~Lu and Y.~Peng, ``A semi-supervised learning algorithm on gaussian mixture
  with automatic model selection,'' \emph{Neural Processing Letters}, vol.~27,
  no.~1, pp. 57--66, 2008.

\bibitem{13722894}
S.~Yang, X.~Wang, L.~Yang, Y.~Han, and L.~Jiao, ``Semi-supervised action
  recognition in video via labeled kernel sparse coding and sparse l1 graph,''
  \emph{Pattern Recognition Letters}, vol.~33, no.~14, pp. 1951 -- 6,
  2012/10/15.

\bibitem{roweis2000nonlinear}
S.~T. Roweis and L.~K. Saul, ``Nonlinear dimensionality reduction by locally
  linear embedding,'' \emph{Science}, vol. 290, no. 5500, pp. 2323--2326, 2000.

\bibitem{lee2006efficient}
H.~Lee, A.~Battle, R.~Raina, and A.~Ng, ``Efficient sparse coding algorithms,''
  in \emph{Advances in neural information processing systems}, 2006, pp.
  801--808.

\bibitem{wang2009linear}
J.~Wang, F.~Wang, C.~Zhang, H.~C. Shen, and L.~Quan, ``Linear neighborhood
  propagation and its applications,'' \emph{Pattern Analysis and Machine
  Intelligence, IEEE Transactions on}, vol.~31, no.~9, pp. 1600--1615, 2009.

\bibitem{Simpson1994342}
E.~Simpson, M.~Mahendroo, G.~Means, M.~Kilgore, M.~Hinshelwood,
  S.~Graham-Lorence, B.~Amarneh, Y.~Ito, C.~Fisher, M.~Michael, C.~Mendelson,
  and S.~Bulun, ``Aromatase cytochrome p450, the enzyme responsible for
  estrogen biosynthesis,'' \emph{Endocrine Reviews}, vol.~15, no.~3, pp.
  342--355, 1994.

\bibitem{BajRossi2014283}
C.~Baj-Rossi, T.~Rezzonico~Jost, A.~Cavallini, F.~Grassi, G.~De~Micheli, and
  S.~Carrara, ``Continuous monitoring of naproxen by a cytochrome p450-based
  electrochemical sensor,'' \emph{Biosensors and Bioelectronics}, vol.~53, pp.
  283--287, 2014.

\bibitem{Rasmussen2014255}
M.~Rasmussen, C.~Klausen, and B.~Ekstrand, ``Regulation of cytochrome p450 mrna
  expression in primary porcine hepatocytes by selected secondary plant
  metabolites from chicory (cichorium intybus l.),'' \emph{Food Chemistry},
  vol. 146, pp. 255--263, 2014.

\bibitem{Guengerich2006E105}
F.~Guengerich, ``Cytochrome p450s and other enzymes in drug metabolism and
  toxicity,'' \emph{AAPS Journal}, vol.~8, no.~1, pp. E105--E111, 2006.

\bibitem{Rostkowski20132051}
M.~Rostkowski, O.~Spjuth, and P.~Rydberg, ``Whichcyp: Prediction of cytochromes
  p450 inhibition,'' \emph{Bioinformatics}, vol.~29, no.~16, pp. 2051--2052,
  2013.

\bibitem{Faulon2008225}
J.-L. Faulon, M.~Misra, S.~Martin, K.~Sale, and R.~Sapra, ``Genome scale enzyme
  - metabolite and drug - target interaction predictions using the signature
  molecular descriptor,'' \emph{Bioinformatics}, vol.~24, no.~2, pp. 225--233,
  2008.

\bibitem{Faulon2003721}
J.-L. Faulon, C.~Churchwell, and D.~Visco~Jr., ``The signature molecular
  descriptor. 2. enumerating molecules from their extended valence sequences,''
  \emph{Journal of Chemical Information and Computer Sciences}, vol.~43, no.~3,
  pp. 721--734, 2003.

\bibitem{Faulon2003707}
J.-L. Faulon, D.~Visco~Jr., and R.~Pophale, ``The signature molecular
  descriptor. 1. using extended valence sequences in qsar and qspr studies,''
  \emph{Journal of Chemical Information and Computer Sciences}, vol.~43, no.~3,
  pp. 707--720, 2003.

\bibitem{Rojatkar20131222}
D.~Rojatkar, K.~Chinchkhede, and G.~Sarate, ``Handwritten devnagari consonants
  recognition using mlpnn with five fold cross validation,'' in
  \emph{Proceedings of IEEE International Conference on Circuit, Power and
  Computing Technologies, ICCPCT 2013}, 2013, pp. 1222--1226.

\bibitem{gao2010local}
S.~Gao, I.~W. Tsang, L.-T. Chia, and P.~Zhao, ``Local features are not
  lonely--laplacian sparse coding for image classification,'' in \emph{Computer
  Vision and Pattern Recognition (CVPR), 2010 IEEE Conference on}.\hskip 1em
  plus 0.5em minus 0.4em\relax IEEE, 2010, pp. 3555--3561.

\bibitem{DePaola2013}
A.~De~Paola, G.~Lo~Re, F.~Milazzo, and M.~Ortolani, ``Qos-aware fault detection
  in wireless sensor networks,'' \emph{International Journal of Distributed
  Sensor Networks}, vol. 2013, 2013.

\bibitem{tian14a}
Y.~Tian, B.~Zhang, E.~P. Hoffman, R.~Clarke, Z.~Zhang, I.-M. Shih, J.~Xuan,
  D.~M. Herrington, and Y.~Wang, ``Kddn: an open-source cytoscape app for
  constructing differential dependency networks with significant rewiring,''
  \emph{Bioinformatics}, 2014.

\bibitem{tian14b}
------, ``Knowledge-fused differential dependency network models for detecting
  significant rewiring in biological networks,'' \emph{BMC Systems Biology},
  vol.~8, p.~87, 2014.

\end{thebibliography}
\end{document}